\documentclass{IEEEtran4PSCC}

\ifCLASSINFOpdf
   \usepackage[pdftex]{graphicx}
\else
   \usepackage[dvips]{graphicx}
\fi

\usepackage[cmex10]{amsmath}
\usepackage[utf8]{inputenc} 
\usepackage[T1]{fontenc}    
\usepackage[hyperfootnotes=false]{hyperref} 
\usepackage{url}            
\usepackage{booktabs}       
\usepackage{amsfonts}       
\usepackage{nicefrac}       
\usepackage{microtype}      
\usepackage{graphicx}
\usepackage{doi}
\usepackage{xcolor}
\usepackage{amsmath}
\usepackage{tikz}
\usepackage{amssymb}
\usepackage{booktabs}
\usepackage{siunitx}
\usepackage{amsmath}
\usepackage{graphicx}
\usepackage{enumitem}
\usepackage{numprint}
\usepackage{caption}
\usepackage{siunitx}
\usepackage{float}
\usepackage{booktabs,multirow,threeparttable,caption,adjustbox}
\usepackage{tabularx}
\usepackage{enumitem}
\usepackage{adjustbox}
\usepackage{tcolorbox}
\usepackage{arydshln}
\usepackage{float} 
\usepackage{stfloats} 
\usepackage{caption}

\usepackage{microtype}

\usepackage[table]{xcolor}
\usepackage[table]{xcolor}
\definecolor{R}{HTML}{C5E0B4} 
\definecolor{C}{HTML}{D9E1F2} 
\definecolor{B}{HTML}{F4CCCC} 

\newcommand{\PC}{PowerChain}  
\newcommand{\PCabbr}{PC}  

\newcommand{\PCO}{PowerChain-O}
\newcommand{\PCOabbr}{PC-O}

\newcommand{\PCM}{Context-Augmented Tool Calling with Descriptor}
\newcommand{\PCMabbr}{C-TC$_\text{D}$}

\newcommand{\ZeroCtx}{Context-Augmented Tool Calling}
\newcommand{\ZeroCtxabbr}{C-TC}

\newcommand{\Workflow}{\textit{workflow}}
\newcommand{\Workflowc}{\textit{Workflow}}

\newcommand{\Tool}{\textit{tool}}
\newcommand{\Tools}{\textit{tools}}
\newcommand{\ToolPool}{\textit{tool set}}
\newcommand{\ToolDesc}{\textit{tool descriptor}}

\newcommand{\KPool}{\textit{knowledge base}}

\newcommand{\AWF}{annotated verified workflow-task pairs}
\newcommand{\AWFc}{Annotated verified workflow-task pairs}

\newcommand{\Task}{\textit{task}}
\newcommand{\prompt}{\textit{prompt}}
\newcommand{\UDB}{\textit{utility database}}

\newcommand{\Verifier}{\textit{verifier}}

\makeatletter
\let\old@ps@headings\ps@headings
\let\old@ps@IEEEtitlepagestyle\ps@IEEEtitlepagestyle
\def\psccfooter#1{%
    \def\ps@headings{%
        \old@ps@headings%
        \def\@oddfoot{\strut\hfill#1\hfill\strut}%
        \def\@evenfoot{\strut\hfill#1\hfill\strut}%
    }%
    \def\ps@IEEEtitlepagestyle{%
        \old@ps@IEEEtitlepagestyle%
        \def\@oddfoot{\strut\hfill#1\hfill\strut}%
        \def\@evenfoot{\strut\hfill#1\hfill\strut}%
    }%
    \ps@headings%
}
\makeatother

\begin{document}

\title{PowerChain: A Verifiable Agentic AI System for Automating Distribution Grid Analyses}

\author{%
  Emmanuel O. Badmus$^1$$^*$, Peng Sang$^1$$^*$, Dimitrios Stamoulis$^2$$^\dagger$, Amritanshu Pandey$^1$$^\dagger$\\
   $^1$Dept. of Electrical
and Biomedical Engineering, 
  The University of Vermont\\
  $^2$Dept. of Electrical and Computer Engineering, 
  The University of Texas at Austin\\
  \texttt{ebadmus@uvm.edu, 
peng.sang@uvm.edu, dstamoulis@utexas.edu, apandey1@uvm.edu}
}

\maketitle

\begin{abstract}
Rapid electrification and decarbonization are increasing the complexity of distribution grid (DG) operation and planning, necessitating advanced computational analyses to ensure reliability and resilience. 
These analyses depend on disparate workflows comprising complex models, function calls, and data pipelines that require substantial expert knowledge and remain difficult to automate. Workforce and budget constraints further limit utilities’ ability to apply such analyses at scale. 
To address this gap, we build an agentic system PowerChain, which is capable of autonomously performing complex grid analyses. 
Existing agentic AI systems are typically developed in a bottom-up manner with customized context for predefined analysis tasks; therefore, they do not generalize to tasks that the agent has never seen. 
In comparison, to generalize to unseen DG analysis tasks, PowerChain dynamically generates structured context by leveraging supervisory signals from self-contained power systems tools (e.g., GridLAB-D) and an optimized set of expert-annotated and verified reasoning trajectories. 
For complex DG tasks defined in natural language, empirical results on real utility data demonstrate that \PC{} achieves up to a $\sim$144\% improvement in performance over baselines.

\thanksto{\noindent Submitted to the 24th Power Systems Computation Conference (PSCC 2026).\\$^*$Equal contribution ~ $^\dagger$Equal last author.}
\end{abstract}

\begin{IEEEkeywords}
Agentic AI, LLMs, Distribution Network
\end{IEEEkeywords}

\section{Introduction} 


\noindent As renewable energy, electric vehicles, and distributed energy resources (DERs) scale up, the utilities that operate and design the electric distribution grids face growing challenges \cite{ray2023review,bank2013analysis}.
Operators must manage variable electricity generation from rooftop solar and batteries, rapid load changes \cite{ray2023review}, and bidirectional power flows \cite{baviskar2020challenges}; planners, on the other hand, need to design networks for higher electrification \cite{crozier2024distribution} and improved resilience.
Increasingly, both operators and planners depend on a wide slew of analyses pipelines to perform their jobs.
For example, a distribution utility engineer may want to evaluate whether a specific capacity rooftop solar PV installation causes voltage or flow violations on a particular feeder.   

The status quo for many of these organizations, especially smaller rural cooperatives and municipalities, is to use a rule of thumb or engineering judgment to perform such analyses, as they may have limited R\&D budgets and a smaller workforce \cite{costello2016primer}.
Without re-envisioning analysis capabilities, the current paradigm is likely to serve as a bottleneck to the growth of electrification and decarbonization efforts on distribution grids.
Unfortunately, the cost of hiring and training experts to perform these analyses is generally intractable for these organizations~\cite{stamoulis2025geo}.
An affordable choice is to automate the execution of distribution grid analyses by using reasoning-based capabilities of large
language models (LLMs) and tool-augmented agents to minimize such costs.

Recent work has demonstrated the potential of LLMs in the power systems domain:
\cite{jia2024enhancing} builds an agentic approach for power system simulation; 
\cite{lin2025novel} invokes language models to estimate distributed solar PV output forecasts; 
\cite{jin2024chatgrid} leverages LLMs for the visualization of power grid operations;
These works, however, rely on well-defined, manually crafted prompts focusing on a specific type of tasks, and are likely to fail against \textit{unseen} distribution network (DN) analyses as task complexity, data pipelines, and reasoning steps increase. For instance, to solve the DN dispatch problem,~\cite{yang2025large} requires predetermined sequences of steps to be explicitly provided as input to the LLM. To our knowledge, \textit{no} existing work builds a general framework for automating distribution grid analysis tasks (or power systems in general) for \textit{unseen} computation tasks.

Advances in agentic AI have evolved LLMs into autonomous agents capable of multi-step reasoning, dynamic tool
invocation, and complex user interactions~\cite{andrews2025scaling}. Based on the novel paradigm of agentic \textit{workflow automation}~\cite{zhang2025aflow}, recent work leverages methods such as Monte Carlo Tree Search (MCTS) or reinforcement learning (RL) to iteratively adjust workflows on coding and question-answering (QA) agentic tasks~\cite{niu2025flow, zhang2025aflow, zhang2025multi,li2024autoflow, hu2025adas}. 
Notably,~\cite{andrews2025scaling, su2025scaling} introduce Agents Research Environments (AREs) featuring dynamic LLM orchestration engines that scale to multi-tool web agents, achieving impressive performance on Wiki QA and mobile application benchmarks. Despite this progress, we observe the following challenges when applying such frameworks to unseen expert-level tasks in the power system domain:

\textit{Limited generalizability}: existing agentic methods for general knowledge benchmarks are built bottom-up~\cite{andrews2025scaling}: human experts are cognizant of how to solve the downstream tasks~\cite{bhattaram2025geoflow} and manually define workflow supervision directly tied to expected LLM actions (e.g., unit-testing correctness prompts in coding benchmarks)~\cite{yang2025large}.
However, extending these systems to unseen tasks remains challenging, as agents must infer new reasoning and tool invocation strategies beyond the scope of their prescribed prompt–action pairs. Therefore, manual workflow specification introduces scalability limits in DN analysis frameworks, where every new computation task requires deep, domain-specific expertise.

\textit{Expert-level verification}: General knowledge tasks like Wiki QA or code uni-testing offer clear pass/fail supervision signals for workflow automation~\cite{zhang2025multi}, where answers such as ``\textit{The capital of France is Paris}'' are atomically verifiable. However, such an ability to \textit{soft}-verify LLM outputs breaks down in DN analysis tasks, where the solutions necessitate expert knowledge and long-horizon reasoning to determine agentic correctness. 

\textit{System costs}: recent analysis in~\cite{su2025scaling} shows that supervised fine-tuning (SFT) might lock agents into replicating fixed behavior patterns rather than developing flexible decision-making, requiring billions of training samples to generalize to unseen web and mobile tasks. Such scaling expenses also persist even without agentic training, as agentic workflow generation and inference through commercial LLMs (e.g., OpenAI's ``Deep Research'' model) incur orders-of-magnitude higher costs than standard function-calling LLMs~\cite{stamoulis2025geo}. Therefore, it is important to enable DN analyses with cost-efficient agentic performance, especially with open-source language models.

To address these gaps, we build \textit{\PC{}}, an agentic workflow automation framework for complex DN analyses. We improve the state-of-the-art with the following key insights: 
(i) we provide a \textit{structured} information-dense expert knowledge context to tool-augmented agents, as \AWF{} exemplars, reducing the supervision cost by orders of magnitude compared to LLM fine-tuning alternatives;
(ii) we \textit{dynamically} optimize the in-context knowledge demonstrations to the agentic system to achieve a trade-off between cost and performance; and iii) we embed an error-context reasoning within the path execution of graph-based workflows to enforce a feedback system with auto-correction.

We represent power-system analyses as executions on generated directed acyclic graphs (DAGs), also referred to as workflows. 
In a workflow DAG, nodes represent tool calls and their associated arguments, while edges capture their dependencies, enabling the agent to interpret user queries, select domain-specific tools, and coordinate their execution. Unlike existing DAG-driven agentic prompting schemes, \PC{} employs \textit{structured}, \textit{verified}, \textit{expert-annotated workflow-task pairs} and a domain-aware knowledge pool to generalizes across DN analyses with real-world distribution data and feeders. Given a user query, the agent does not just recall or match seen tasks, but iteratively composes (\textit{orchestrates}) and tests (\textit{verifies}) multi-step workflows until convergence and task completion. 

To our knowledge, this is the \textit{first} agentic system to automate expert-level DN analysis workflows, enabling robust and timely grid assessments without prescribed complex scripting or rigid pipeline reconfigurations.
To demonstrate empirical performance, we evaluate our proposed framework against various analysis tasks on real-world distribution and feeder data. The main contributions of this work are as follows:
\begin{itemize}[nosep, leftmargin=5.5mm]
    \item \textit{\PC{} generates workflows to execute unseen analysis tasks} using in-context learning by enabling LLMs to leverage domain-aware tool descriptors and expert-built, annotated, verified workflows. 

    \item Compared to unstructured RAG baselines (see Section~\ref{sec:modes}), \textit{\PC{}} increases pass@1 in the absolute range of 0.04-0.29, corresponding to roughly a 10\%–60\% relative gain; increases precision in the absolute range 0.07-0.47, corresponding to a 30\%–200\% relative gain. 
    \textit{\PC{}} shows the best performance across all LLMs for all tasks.
     
    \item \textit{\PC{} democratizes model-based distribution grid analyses} by (i) requiring no LLM fine-tuning on domain-aware workflows and (ii) by being locally deployable on lightweight open-source models (e.g., Qwen3 8b).

    \item \textit{\PC{} utilizes optimal \AWF{} subset selection} to achieve better accuracy with fewer inputs while also reducing token cost.
\end{itemize}


\section{\PC{} Foundations}
\label{sec:terminology}
\noindent Our \PC{} \textit{environment} ($\mathcal{E}_{\text{PC}}$) comprises the agents, the agentic tool set, the annotated verified workflow-task pairs, and the utility data which together form the core of our agentic system. \textit{Agent} ($\mathcal{A}$) is a tool-augmented LLM that, given a user query, acts as the reasoning and decision-making model to complete complex tasks, in our case, distribution grid analyses. 
In \PC{}, the agent has two distinct \textit{roles}: \textit{orchestrator} ($\mathcal{O}$) and \textit{verifier} ($\mathcal{V}$).
The orchestrator ($\mathcal{O}$) leverages structured, expert-annotated workflow-task pairs to iteratively generate a candidate workflow. The verifier ($\mathcal{V}$) executes and validates the candidate workflow by the orchestrator.

A \textit{workflow} is a directed acyclic graph (DAG) with a tool and respective arguments at each node.
Executing the \textit{correct} workflow solves the distribution grid compute task.
\textit{Tools} ($t$) are functions that are exposed to the agent as standalone APIs. 
The set of all available tools represents the \textit{tool set}.
The \textit{verified workflow-task pairs} capture expert-annotated domain knowledge to assist with workflow generation.
See Section~\ref{sec:PC_design} for comprehensive design of these elements in \PC{}.

\section{Related Work}
\noindent \textbf{Problem Statement:} \textit{Can we develop a tool-augmented agentic system that generates stateful DAG workflows with a tool and relevant arguments at each node to successfully execute an unseen distribution grid analysis task?} Given the rapid progress of large language models (LLMs) and agentic AI, it is important to systematize and review the current \textit{frontier} methods that could serve as building blocks to address the aforementioned challenge from the perspective of a power grid analyst. 

\subsection{Tool-augmented LLMs}
\label{sec:agents}
\noindent \textbf{Baseline agentic LLMs.} Agents represent one of the most significant developments in language models, extending their capabilities beyond text generation to interaction with external tools and APIs~\cite{yang2025qwen3, openai2025o3}. In the power systems domain, Yang~\cite{yang2025large} demonstrated the use of an agent to solve the DN dispatch problem. However, to achieve robust reasoning about task sequencing~\cite{singh2024geollm}, such approaches
often rely on predefined execution \textit{templates}, where task objectives were explicitly broken down in the LLM prompts using structured pragma-style \textit{tags} (e.g., ``\textit{Prompt}: <\texttt{objective}> perform analysis </\texttt{objective}> ... <\texttt{input}> \texttt{filename} </\texttt{input}>''~\cite{yang2025large, singh2024geollm}). Such rigid prompting requires designing separate templates for each task type, constraining generalization and scalability beyond the prescribed sequence of steps.

\textbf{Agentic workflows.} 
Despite their impressive capabilities, baseline agents often struggle with complex multi-step tasks that require sophisticated reasoning, especially due to the reliance on implicit self-feedback for error corrections and tool usage~\cite{su2025scaling}.
Recent studies~\cite{li2024autoflow, andrews2025scaling, bhattaram2025geoflow, li2025chain, fang2025memp} show that agentic performance improves when models incorporate annotated, reusable experiences derived from prior model executions or human-annotated trajectories; such in-context LLM supervision has been variously referred to in prior work as \textit{memories}~\cite{fang2025memp}, \textit{workflows}~\cite{bhattaram2025geoflow}, \textit{scenarios}~\cite{andrews2025scaling}, or \textit{knowledge}~\cite{su2025scaling} traces, or \textit{trajectories}~\cite{li2025chain}. However, constructing and maintaining such expert-level \textit{experience pool} introduces trade-offs between scalability and correctability~\cite{su2025scaling}.

\subsection{Enhancing Agents via RAG and fine-tuning}

\noindent \textbf{Retrieval-Augmented Generation.}
RAG is a widely adopted approach for augmenting LLMs with external knowledge~\cite{balaguer2024rag}. Information from various data sources is embedded into a vector store~\cite{faiss2019}, against which the model performs a similarity search to retrieve relevant content. ~\cite{jia2024enhancing} adapts RAG for power system analysis to augment a feedback-driven framework with simulation APIs on 69 diverse tasks from \texttt{DALINE}~\cite{jia2024user} and \texttt{MATPOWER}\cite{matpower2024manual}. 
RAG's effectiveness is however limited by its reliance on semantic similarity between the query and stored context~\cite{misrahi2025adapting}. As also shown in our Results (Section~\ref{sec:PerfComp}), retrieval from unstructured data that appears textually relevant but lacks actionable information performs poorly on complex, multi-step reasoning DN tasks~\cite{zeng2025worse}.

\textbf{Supervised fine-tuning.}
SFT remains the standard approach for adapting AI models to domain-specific tasks~\cite{li2024autoflow}. In the power systems domain,~\cite{cheng2025large} tailors an LLM for power dispatch tasks formulated as multiple-choice, single-choice, and QA questions on basic concepts. However, unlike post-training on such simpler ChatGPT-style text-completion queries, agentic SFT (and, RL-based fine-tuning in particular~\cite{gao2025beyond, openai2025deepresearch}) requires fine-tuning on multi-step trajectory data spanning a vast reasoning space~\cite{su2025scaling}. Despite strong performance on general-purpose benchmarks (e.g., web and mobile applications~\cite{li2025chain, gao2025beyond}), recent work demonstrates that SFT effectiveness often diminishes in expert-level domains, locking agents into fixed behavioral patterns rather than developing generalizable decision-making capabilities~\cite{su2025scaling}. Addressing this limitation demands training on massive numbers of synthetic workflow permutations (up to 315 billion reported for web agents~\cite{su2025scaling}), placing meaningful SFT on DN problems out of reach for smaller research groups, utilities, or public institutions~\cite{stamoulis2025geo}.

\subsection{Workflow Automation and Agentic Environments}

\noindent \textbf{Agentic workflow automation.}
Training-free workflow automation harnesses the reasoning benefits of structured, \textit{annotated}, \textit{in-context} demonstrations while reducing supervision requirements to only tens or hundreds of workflow examples -- several orders of magnitude fewer than those required for SFT. Recent works~\cite{niu2025flow,zhang2025aflow,li2024autoflow,bhattaram2025geoflow,zhang2025multi,hu2025adas} explore lightweight traversal of the agentic trajectory space using MCTS, RL, or regression-based heuristics. While we build upon this paradigm, existing methods primarily target general-knowledge domains such as web search, coding, and Wiki QA~\cite{andrews2025scaling}, where reasoning can be formulated bottom-up: human experts explicitly define workflow supervision aligned with expected LLM actions (e.g., unit-testing prompts in coding tasks)~\cite{bhattaram2025geoflow, yang2025large}. To our knowledge, no prior workflow automation techniques address power grid analysis, and as shown in our results, applying off-the-shelf methods to DN tasks often fails to generalize to \textit{unseen} scenarios (Section~\ref{sec:PerfComp}).

\textbf{State-of-the-art Agentic Environments.} Real-world expert-level tasks require agents to sustain complex, long-horizon interactions with realistic self-contained APIs and external codebases~\cite{andrews2025scaling}. However, most existing agentic automation approaches assume idealized models of agent operation~\cite{jimenezswe, yao2024tau}, resting on two major limitations. First, tools are often \textit{custom-built} for the agent (rather than the opposite, i.e., designing agents that can operate with existing APIs), implicitly shaping the workflows they are meant to evaluate~\cite{singh2024geollm, yang2025large, niu2025flow}. Second, current evaluation metrics often reduce agentic performance to final-state validation, checking only whether the function and argument strings match the target tool calls, without further assessing the correctness or coherence of intermediate reasoning steps~\cite{patil2025bfcl}.

Recent large-scale evaluations from industry labs~\cite{su2025scaling, andrews2025scaling, li2025chain} highlight two main requirements for scaling agentic systems to real-world applications. First, rather than relying on bespoke tool sets, agents must support \textit{generalizable} tool sets and flexible integration of external APIs, primarily achieved through compatibility with the industry-standard  Model Context Protocol (MCP). Second, evaluation must move beyond endpoint accuracy toward \textit{full-path} assessment~\cite{michelakis2025core} across every step of the reasoning trajectory. As detailed next (Sections~\ref{sec:method}-\ref{sec:expsetup}), our framework is the \textit{first} agentic environment for DN tasks to adopt both principles: we implement an MCP-based backend that automatically populates tools from validated analytical codebases~\cite{foster2022three, pandey2018robust, badmus2024anoca, panthee2025solving}, and we evaluate performance using agentic full-path correctness and precision metrics~\cite{michelakis2025core, andrews2025scaling} that account for all actions across the DAG workflow.

\section{Methodology}
\label{sec:method}

\noindent \PC{} automates distribution grid analyses by assimilating foundational elements in Section~\ref{sec:terminology}. It inputs a user task $q$ in natural language, generates a workflow $w$ as a directed acyclic graph (DAG) of tools and argument pairs, and executes it with data from the utility database ($D$) to perform the underlying compute task $q$. 
\PC{} is feedback-based. It can autocorrect its workflows until a verifiably correct workflow emerges. 
\PC{} can generate workflows to autonomously solve \textit{unseen} distribution grid tasks without human-in-the-loop by reasoning from annotated and verified expert knowledge and tool descriptors, resulting in a generalized agentic framework that operates with both open and proprietary LLM models.




\subsection{\PC{} Overview}\label{sec:PC_overview}



\begin{figure*}[ht!]
    \centering
    \includegraphics[width=1.0\linewidth]{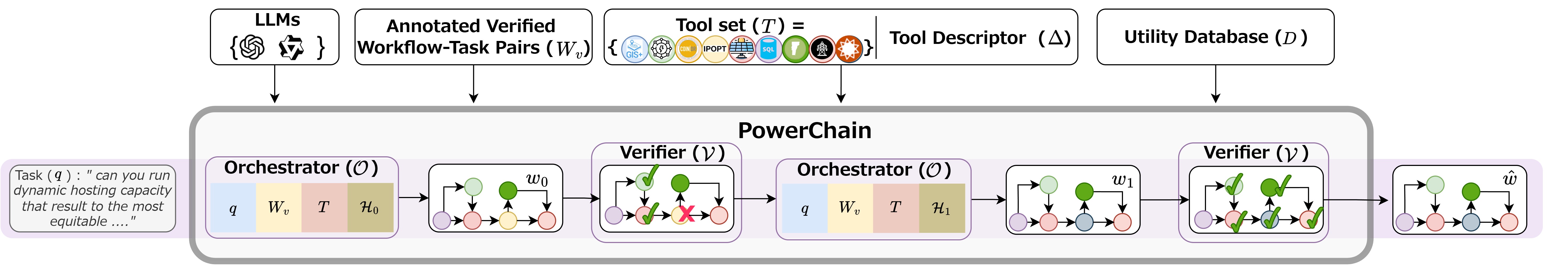}
    \caption{ \small \PC{} agentic \Workflow{} generation framework. User tasks are inputs to the orchestrator ($\mathcal{O}$), which builds prompts using \AWF{} ($W_v$), \ToolDesc{} ($\Delta$), and conversation history ($\mathcal{H}$). An LLM generates candidate \Workflow{} $w$ that relate elements in the \ToolPool{} ($T$) and the \UDB{} ($D$). The \Verifier{} ($\mathcal{V}$) tests \Workflow{}, with errors fed back to the \Verifier{} via an LLM until a valid \Workflow{} $\hat{w}$ is produced.}
    \label{fig:PowerChain_flowchart}
\end{figure*}

\noindent \PC{} builds an agentic architecture (Fig.~\ref{fig:PowerChain_flowchart})  that leverages LLM, \AWF{} ($W_v$), and tool descriptors ($\Delta$)  in the power systems \ToolPool{} ($T$) to dynamically \textbf{build-verify-feedback-correct} workflows $w$ to perform a task given in an unseen user query $q$.
The agentic framework is recursive. The agent orchestrator $\mathcal{O}$ generates a DAG with tool-argument pairs ($w$) at its nodes, namely workflow, by constructing a domain-aware dynamic prompt $\mathcal{P}$ and invoking an LLM.
The \PC{} environment has access to tools in \ToolPool{} $T$ through API calls for functionalities such as fetching data from the utility database $(D)$, parsing network data, and running power flows.
The agent verifier $\mathcal{V}$ executes the workflow and decides whether the given workflow is correct or erroneous.
Upon receiving an error, the verifier provides the orchestrator with feedback regarding the error context.
The orchestrator re-invokes the LLM with the error context to generate a revised workflow.
The cycle repeats until the agent generates an optimal workflow $\hat{w}$ or a prespecified counter is hit.
\begin{align}
    \hat{w} &= \textbf{\PC}\!\big\{\,q, W_v,\,T,\,\Delta,\,D,\,\text{LLM}\,\big\}
    \label{eq:powerchain}
\end{align}

\subsection{\PC{} Design}\label{sec:PC_design}
\noindent Consider a user task $q$: 
\textit{``Conduct an L1-norm-based three-phase AC power flow infeasibility check for the South Hero feeder in Vermont at 10:00 AM on March 22, 2025. Use solar and load data at that time to initialize the model, solve the optimization problem to check whether the power flow model is feasible, and list all buses that violate the voltage bounds."}\footnote{Check \cite{foster2022three} for details on three-phase power flow infeasibility}

The orchestrator will dynamically construct a prompt $\mathcal{P}$ by collating the user $q$ with $W_v$, conversation history ($\mathcal{H}$), and \Tool{} descriptors ($\Delta$).
We define the prompt at the step $k$ in \eqref{eq:prompt_def}.
\begin{equation}
    \mathcal{P}_k = \{q,W_v,\mathcal{H}_k,\Delta\} \label{eq:prompt_def}
\end{equation}
The orchestrator passes the prompt $\mathcal{P}_k$ to the LLM, which generates a workflow $w_k$ as shown in \eqref{eq:wk}.
\begin{equation}
    w_k = \text{LLM}(\mathcal{P}_k)\label{eq:wk}
\end{equation}
At each step $k$, the agent verifier $\mathcal{V}$ evaluates the workflow $w_k$ from the orchestrator.
Using the power systems \ToolPool{} $T$ including tools for accessing the datasets from the \UDB{} $D$, the agent executes the \Workflow{} as an ordered sequence (from DAG's path) of \Tools{} and their respective \textit{arguments}.

The \Tools{} return error context as exit strings from the execution at $k^{th}$ step,  which are stored in the list $Y_k = [y_k^1, y_k^2,...,y_k^m]$, where $m$ is the number of tool calls in the workflow $w_k$.
The orchestrator augments the conversation history to include this information for the next step. 
\begin{equation}
    \mathcal{H}_{k+1} = \mathcal{H}_k \oplus \big(w_k, Y_k\big)
\end{equation}
Next, the orchestrator constructs the prompt $\mathcal{P}_{k+1}$ to invoke an LLM.
If LLM execution is verified (i.e., successful without errors), the resulting workflow $w_k$ is deemed complete and returned as $\hat{w}$.
Otherwise, the LLM produces a revised workflow $w_{k+1}$ and repeats the process of \textit{build-verify-feedback-correct} until a valid $\hat{w}$ is obtained. Obtaining the optimal outcome, in theory, would correspond to matching the expert workflow. However, note that in practice, expert workflows are unavailable for \textit{unseen} tasks. 
Verification is done through error checks of the tool.
See the summary of this design in Fig.~\ref{fig:PowerChain_flowchart}.

\subsubsection{\Workflowc{} ($w$)}
A workflow is an LLM-generated DAG, where nodes $V$ include \Tools{} and their arguments and graph connections $C$ represent their dependencies. 
Formally, we represent a workflow of length $m$ as:
\begin{equation}
w = (V, C), \qquad 
V = \{(t_i, \Theta_i)\}_{i=1}^{m}, \quad 
C \subseteq [m] \times [m]\label{eq:workflow_m}
\end{equation}
Each pair $({t}_i, {\Theta}_i)$ denotes a tool invocation $t_i$ with its argument set $\Theta_i$. 
For the example query in Section~\ref{sec:PC_design}, the workflow execution as a sequence of steps is shown in Fig.~\ref{fig:example_workflow}.

\begin{figure}[htpb]
\centering
\begin{tcolorbox}[colback=gray!5, colframe=gray!50, boxrule=0.5pt, arc=3pt, left=4pt, right=4pt, top=2pt, bottom=2pt]
\begin{adjustbox}{scale=0.75}
\begin{minipage}{\linewidth}
\[
\begin{aligned}
&\textbf{\PC{} Environment }(\mathcal{E}_{\text{PC}})\ \textbf{— REGISTRY} \\
&\quad \texttt{data = \{\}} \\
&\quad \texttt{model = \{\}} \\
& \\
&\textbf{WORKFLOW} \\
&\texttt{\textbf{fetch\_AMI}(time="2025-03-22 10:00",} \\
&\qquad \texttt{feeder="south\_hero", reg\_out="data")} \\
\to\;&\texttt{\textbf{load\_solar}(time="2025-03-22 10:00", lat=44.65,} \\
&\qquad \texttt{lon=-73.3, reg\_out="data")} \\
\to\;&\texttt{\textbf{load\_network}(feeder="south\_hero", reg\_out="data")} \\
\to\;&\texttt{\textbf{init\_model}(reg\_in="data", reg\_out="model")} \\
\to\;&\texttt{\textbf{add\_constraints}(type="dhc\_L1", reg\_in="model")} \\
\to\;&\texttt{\textbf{add\_objective}(type="dhc\_L1", reg\_in="model")} \\
\to\;&\texttt{\textbf{solve\_model}(reg\_in="model")} \\
\to\;&\texttt{\textbf{update\_voltages}(reg\_in="model", reg\_out="data")} \\
\to\;&\texttt{\textbf{check\_voltage\_violations}(reg\_in="data")}
\end{aligned}
\]
\end{minipage}
\end{adjustbox}
\end{tcolorbox}
\caption{Example of a correct workflow $w$ for the query in Section~\ref{sec:PC_design}. 
The workflow shows the ordered tool sequence that successfully executes the task without errors.}
\label{fig:example_workflow}
\end{figure}

\begin{figure*}[htbp]
    \centering
    \includegraphics[width=0.7\linewidth]{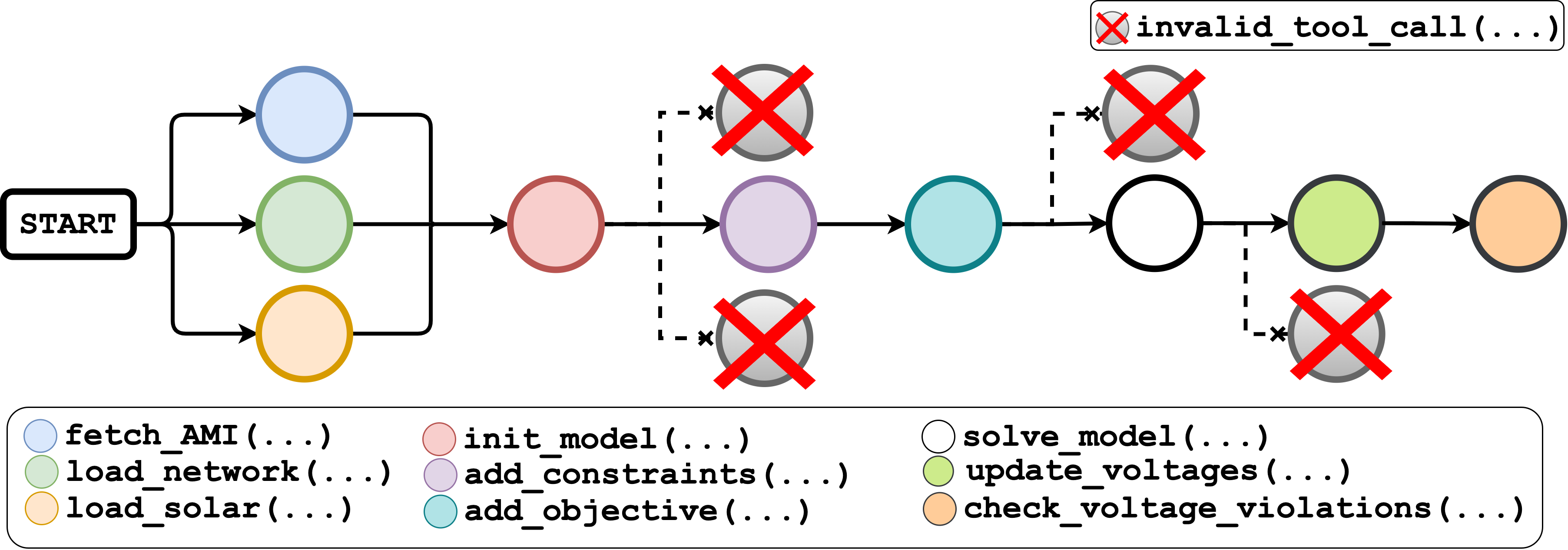}
    \caption{
    Workflow correction trace for the example query in Section~\ref{sec:PC_design}. 
    The DAG shows several iterations of \PC{} during the build–verify–feedback–correct process. 
    Nodes marked “x” correspond to invalid tool calls detected in earlier iterations, while the connected valid nodes represent the final corrected workflow path.}
    \label{fig:placeholder}
\end{figure*}



\subsubsection{Tool descriptor ($\Delta$)} 
For a \Tool{} $t$ in \ToolPool{} $T$, the \Tool{} \textit{descriptor} is defined as $\delta= (\phi, \Theta, d)$ and the set of all descriptors is defined as $\Delta$.
Each \Tool{} descriptor $i$ contains a unique \Tool{} name ($\phi_i$), a set of arguments with $n$ individual arguments ($\Theta_i = \{ \theta^k_i \mid k=1,\dots,n$ \}), and an optional text description of the \Tool{} ($d_i$).

\subsubsection{\AWFc{} ($W_v$)} PowerChain uses a curated set of expert workflow-query pairs to guide in-context reasoning:
\begin{equation}
W_v = \{(w_v,q_v)|w_v \in \mathcal{W}_v,q_v \in Q_v\}
\end{equation}
where each pair $(w_v, q_v)$ links a natural-language query $q_v$ to a annotated verified expert workflow $w_v$.
Each annotated verified workflow task pair represents a realistic analytical task that the available \ToolPool{}{} $T$ can solve. 
A domain expert defines the problem statement, constructs the corresponding executable workflow, and verifies that its outputs match the expected analytical results. Each $(w_v, q_v)$ is reviewed to ensure deterministic behavior. 
These verified pairs demonstrate how experts translate domain questions into structured computational procedures. 
By embedding them in the \prompt{} $\mathcal{P}$, \PC{} enables the LLM to infer correct tool orderings, argument mappings, and reasoning patterns for unseen tasks.

\subsection{Subset Selection of Annotated Verified Workflow-Task Pairs}
\label{sec:optimal_awf_selection}

\noindent Using \AWF{} enables \PC{} to ground its reasoning in verified domain exemplars. However, as the number and diversity of available \AWF{} grow, the prompt can become large and redundant. Not all pairs contribute equally to a given \Task{}, and irrelevant examples may introduce noise that weakens tool selection and reasoning~\cite{sclar2023quantifying}. 
Moreover, long prompts increase token cost since LLM inference scales with input length, slow response time, and may reduce accuracy due to context effects~\cite{TRIPATHI2025841}. 
To maintain precision and efficiency, \PC{} incorporates an optimal subset selection process to identify only the most relevant $W_v$ for each query.


To address these issues, we first \emph{expand} both the user query and each \AWF{} into simplified, semantically clear sentences using an LLM. 
This expansion step serves two purposes: (i) it normalizes differences in phrasing, abbreviations, or domain-specific jargon that could otherwise reduce similarity scores, and (ii) it makes the intent of both the unseen query and the expert queries explicit, thereby aligning them in a common semantic space. 
Once expanded, these queries are less sensitive to surface-level variations and more directly comparable.


We compute the similarity between the expanded user task $q$ and every expanded expert task in the annotated set ($q_v \in Q_v$) with embeddings of the expanded text using the sentence transformer embedding model in~\cite{wang2022text}.
Given the set of $Q_v$ and an application of the embedding model $M_{\text{exp}}(\cdot)\in\mathbb{R}^d$  to the expanded tasks, the similarity for user task $q$ with task $q_v^i$ in annotated set $Q_v$ is defined as:

\begin{equation}
\begin{aligned}
\operatorname{sim}(q,q_v^i)&=\cos\!\left(M_{\text{exp}}(q),M_{\text{exp}}(q_v^i)\right) \\
&=\frac{M_{\text{exp}}(q)\cdot M_{\text{exp}}(q_v^i)}
{\|M_{\text{exp}}(q)\|\,\|M_{\text{exp}}(q_v^i)\|}
\end{aligned}
\label{eq:cosine_similarity}
\end{equation}

Once we compute similarity scores for all annotated tasks $q_v \in Q_v$ with user task $q$, we rank them from highest to lowest scores and store them in $\hat{Q}_v$.
We generate top-$X$ annotated workflow task pairs from the subset:
\begin{equation}
\begin{aligned}
W_v^{\text{sub}} 
= \{(w_i, q_i) \mid (w_i, q_i) \in W_v,\; q_i \in \operatorname{Top}_X(\hat{Q}_v)\}
\end{aligned}
\label{eq:wv_subset}
\end{equation}
The subset $W_v^{\text{sub}}$ replaces $W_v$ in~\eqref{eq:powerchain} for PowerChain-O model: 
\begin{align}
    \hat{w} &= \textbf{PowerChain-O}_X \!\big\{\,q, W^{\text{sub}}_v,\,T,\,\Delta,\,D,\,\text{LLM}\,\big\}
    \label{eq:powerchainx}
\end{align}
where $X$ represents the length of subset $W_v^{sub}$.

\section{Experimental Setup}
\label{sec:expsetup}

\subsection{Network Model and Data Tools}
\label{sec:benchmark_scope}
\noindent We evaluate \PC{} on \textit{real-world distribution feeders} from Vermont Electric Cooperative (VEC). 
A tool, \texttt{fetch\_ami(args)}, connects directly to the VEC AMI server through a secure API. This tool dynamically fetches relevant consumption data for each meter, parses time-series load curves, and formats them for downstream simulation and optimization, given input arguments for date, time, and region. 
Similarly, tool \texttt{load\_solar(args)} estimates solar generation for each PV system based on date and time, geolocation, irradiance data, and installed capacity; and tool \texttt{load\_network(args)} loads and parses the input network data from \texttt{GeoJSON} and \texttt{GridLAB-D} files into a unified Python object for analysis. 
These tools ensure the benchmark operates on accurate utility-sourced network topology, consumption, and generation data.

\subsection{Analysis Types and Tools}
\label{sec:analysis_tools}
\noindent We focus on three analyses for benchmarking \PC{}:

\begin{enumerate}[leftmargin=23pt]
    \item \textit{Steady-State Three-Phase Power Flow}~\cite{pandey2018robust}, which computes bus voltages, branch currents, and power flows under unbalanced network loading.
    \item \textit{Dynamic Hosting Capacity (DHC)} with $\ell_1$, $\ell_2$, and $\ell_{\infty}$ curtailment formulations~\cite{badmus2024anoca, liu2022using, moring2023inexactness, yi2022fair}, estimates the maximum distributed generation that the network can support while maintaining voltage and current limits.
    \item \textit{Current Infeasibility Analysis}~\cite{foster2022three,panthee2025solving}, identifies whether unbalanced three-phase AC power flow constraints hold for the network under study.
\end{enumerate}

To execute these analyses within \PC{}, we employ tools derived from existing, validated repositories that implement the core methods in \cite{foster2022three, pandey2018robust,badmus2024anoca, panthee2025solving}. They provide standard implementations for three-phase power flow, dynamic hosting capacity, and current infeasibility analysis. Each repository exposes solver routines and optimization modules (as Python API classes) commonly used in distribution-grid research and utility studies. 

We implement an MCP-like backend that automatically populates the tool set and interfaces repositories within the PowerChain environment, without the need for custom tool definitions. Together, the analysis-based tools, the data loading tools (Section \ref{sec:benchmark_scope}), and a custom set of tools for plotting and post-processing form a \ToolPool{} of 21 callable tools exposed to the \PC{} environment.

\subsection{Tasks and Annotated Workflow Curation}
\label{sec:task_workflow_curation}

\noindent We curate 40 representative tasks that reflect practical distribution grid computational analyses utility engineers want to solve in the \PC{} environment (see tasks in Appendices~\ref{appx:queries}–\ref{appx:awf}).
The tasks span different difficulty levels across the three analysis families in Section~\ref{sec:analysis_tools}. 
They range from simple tasks that require calling one or two tools, such as counting network components or computing total feeder load, to tasks that require a complex chain of tool calls. 

\textbf{Workflow "quality assurance".} To ensure we design and validate \textit{representative} workflows for each task, we emulate the annotation process proposed in industry-grade agentic engines~\cite{andrews2025scaling}: three domain-experts \textit{independently} designed "oracle" workflows for each task; starting with the first two annotators, if their suggested solutions diverge, we check for prompt ambiguity; if needed, they update their individual solutions given the refined prompt. The third annotator compares the two "oracle" solutions and confirms that they are consistent. If not, they can either reject the task altogether or make minor edits to the best workflow. Last, the annotators verified that no two workflows share identical tool–argument chain solutions, implying task repetition. 

\textbf{Workflow "dataset" splits.} From these 40 verified pairs, 10 were selected to form the annotated verified workflow set (\AWF{}), denoted $W_v$, which also spans the full difficulty range.
The remaining 30 (denoted $W_e \subseteq \mathcal{W}_e \times Q_e$) were held out for evaluation, where $\mathcal{W}_e$ is the set of annotated workflows for evaluation. 
This structure allows \PC{} to be tested on \textit{unseen} but related tasks while maintaining strict separation between annotated workflows exposed to LLMs and evaluation workflows. 
For RAG-related experiments, the \KPool{} consists of the original research papers corresponding to these repositories~\cite{foster2022three,pandey2018robust,badmus2024anoca,panthee2025solving}, ensuring that retrieval inputs are aligned directly with the implemented analytical methods.

\subsection{Agentic Systems (AS)}
\label{sec:modes}
\noindent We evaluate five agentic systems for workflow automation:

\begin{enumerate}[leftmargin=23pt]
\item \textit{\ZeroCtx}  \textit{(\ZeroCtxabbr):} The agent receives the user query $q$ and the available tools defined only by function names and arguments without textual descriptions or \AWF{}.

\item \textit{\PCM}   \textit{(\PCMabbr):} The agent receives $q$ along with full tool descriptors $\Delta$, which include tool names, arguments, and textual descriptions $d$ describing their behavior.
\item \textit{\PCMabbr+RAG:} Same as  \PCMabbr{}, but adds retrieved text from the \KPool{} through RAG. The RAG pipeline builds a searchable index of research papers as a \texttt{FAISS} vector store~\cite{faiss2019} from \texttt{LangChain}~\cite{langchain2024}. Each document is encoded with OpenAI text embeddings~\cite{openai2024embeddings}. The system appends the retrieved text directly to the model prompt before reasoning.

\item \textit{\PC}  \textit{(\PCabbr):} The agent receives $q$, full descriptors $\Delta$, and a set of expert-generated \AWF{} $W_v$. This represents 10 in this case.

\item \textit{\PCO$_X$ (\PCOabbr$_X$):} The agent receives $q$, full descriptors $\Delta$, and only the top-$X$ expert \AWF{} chosen through the process described in Section~\ref{sec:optimal_awf_selection}.

\end{enumerate}

\noindent All systems share the same execution environment, tool API, and scoring protocol. The difference is only in contextual conditioning, which refers to the amount and type of prior structured or unstructured knowledge the model receives before reasoning and execution.

\subsection{Large Language Models (LLMs)}
\label{sec:models}
\noindent The LLMs fall into two categories: i)
\textit{General-purpose proprietary models} from OpenAI: GPT-4o-mini, GPT-4o, GPT-5-mini, and GPT-5; and \textit{reasoning-optimized proprietary models}: o4-mini and o3. 
All are accessed through the OpenAI API and run on a local PC (2.7GHz Intel Core i7, 16 GB RAM), ii) \textit{Open-source models:} Qwen-2.5-7B and Qwen-3-8B. Both are accessed through the Ollama API. These models were run on the GPU server with an NVIDIA Tesla~V100-SXM2 GPU (16~GB VRAM).
We consider different LLMs to emphasize the importance of evaluating models independently of the systems above, as different models specialize in different tasks.
For example, the frontier models like the o3 model are optimized for complex reasoning tasks, whereas previous generation models are optimized for function calling.


\subsection{Evaluation Metrics}
\label{sec:metrics}
\noindent We evaluate agent workflows $\hat{w}$ against expert workflows $w_e \subseteq \mathcal{W}_e$ using two evaluation classes. 
The first checks if the numerical results from the agent match those from the expert workflow. 
The second measures whether the agent used the most optimal sequence of tools to reach that result, as this affects computational efficiency and token usage. 
We adopt established metrics from the agentic AI community~\cite{singh2024geollm,koh2024visualwebarena,zhou2024webarena,chen2021evaluating}, including precision and pass@k, which quantify correctness and efficiency across multiple generations. 

\subsubsection{Correctness Metrics}

\noindent Correctness measures how closely the agent’s workflow matches expert execution.
We evaluate two criteria: i) whether the agentic workflow reaches the same numerical result as the expert workflow, and ii) whether the sequence of tool calls is the same as that of the expert.


\paragraph{Pass@k} 
pass@k measures the probability that at least one of $k$ generated workflows produces the correct numerical result~\cite{chen2021evaluating}. 
Let $c$ be the number of successful runs out of $n$ total attempts for a query. 
The metric is computed as
\begin{equation}
    \text{pass@k} = 1 - \frac{\binom{n - c}{k}}{\binom{n}{k}}, \quad 1 \le k \le n
\end{equation}
A higher pass@k indicates a greater likelihood that the agent finds a correct workflow within fewer attempts.
For $k=1$, pass@1 is the rate of success, which is:
\begin{equation}
    \text{pass@1} 
              = \frac{1}{n}\sum_{i=1}^{n} \mathbf{1}\!\left[R(\hat{w}_i) \equiv R(w_e)\right]
\end{equation}
where $R(\cdot)$ denotes the workflow execution result and $\mathbf{1}[\cdot]$ is the indicator function.

\paragraph{Precision (Pr)} 
Precision quantifies whether the sequence of tool calls between the expert and the agent is equivalent. 
For each attempt $i$, $\text{Pr}_i = 1$ if $\hat{w}_i \equiv w_e$, and $\text{Pr}_i = 0$ otherwise. 
The precision rate is
\begin{equation}
    \text{Pr} = \frac{1}{n}\sum_{i=1}^{n}\text{Pr}_i 
              = \frac{1}{n}\sum_{i=1}^{n} \mathbf{1}\!\left[\hat{w}_i = w_e\right]
\end{equation}

\subsubsection{Efficiency Metrics}
\noindent Efficiency measures the resource cost it takes for a model to generate a correct workflow. Even when two models achieve similar correctness, one may consume more tokens to reach the same result. 
We quantify this cost using a single metric: \textbf{Tokens per pass@1 (Tk/P@1)} as it measures the average number of tokens consumed per successful run:
\begin{equation}
\text{Tk/P@1} =
\begin{cases}
    \dfrac{\sum_{i=1}^{n} T_i}{\sum_{i=1}^{n} \text{pass@1}_i}, & \text{if pass@1} > 0 \\
    -, & \text{if pass@1} = 0
\end{cases}
\end{equation}
where $T_i$ is the total token count (prompt + completion) used in attempt $i$.


Following established agentic evaluation schemes~\cite{andrews2025scaling}, we execute each task \textit{five independent times} per model-mode combination to account for stochastic variation in agent workflow generation; all reported metrics represent the average across these five runs.
\section{Results}
\label{sec:results}
\noindent We evaluate all agentic systems in Section~\ref{sec:modes} across all LLMs in Section~\ref{sec:models}.
We evaluate 30 tasks, and none exist in the \textit{annotated verified workflow-task pairs}.

\subsection{Performance  Comparison across Agentic Systems and LLMs}\label{sec:PerfComp}

\noindent We assess performance with correctness and efficiency metrics in Section~\ref{sec:metrics}.
For \PCOabbr$_X$, we set $X{=}5$. 
We include ten \AWF{}. 


\ZeroCtxabbr{} shows moderate performance on \textit{simple tasks} across proprietary models. These tasks need little reasoning, and the agent reaches correct results in most cases ($\text{pass@1}{=}0.70$–$0.84$) by repeating \textbf{build-verify-feedback-correct} cycles. For example, GPT-4o-mini attains $\text{pass@1}{=}0.70$ but $Pr{=}0.00$, showing that it finds the right tool sequence through retries and self-correction rather than precise reasoning. \PCMabbr{} improves over \ZeroCtxabbr{}, reaching $\text{pass@1}{=}0.90$–$1.00$  with a reasonable $Pr{=}0.70$–$0.78$. The gain comes from tool descriptions that guide tool calls and argument selection. \PCMabbr+RAG{} adds only small and inconsistent gains ($\Delta\text{Pass@1}{<}0.05$, $\Delta\text{Pr}{<}0.05$). \PCOabbr$_X$ and \PCabbr{} achieve near-perfect performance ($\text{pass@1}{\approx}1.00$, $Pr{\geq}0.90$), confirming that structured workflow context ensures stable multi-tool reasoning.
The open-source Qwen models follow the same trend but perform worse overall ($\text{pass@1}{<}0.90$, $Pr{<}0.70$) due to smaller model capacity. Their accuracy improves under \PCOabbr$_X$ and \PCabbr{}, yet they remain below proprietary models across all systems (Table~\ref{tab:avg_all}).

\begin{table}[htpb]
\centering
\footnotesize
\setlength{\tabcolsep}{4pt}
\renewcommand{\arraystretch}{1.05}
\caption{Average performance on all prompts per task group.}
\label{tab:avg_all}
\resizebox{\linewidth}{!}{
\begin{tabular}{@{}ll|cc|cc|cc|cc|cc@{}}
\toprule
\multirow{2}{*}{\rotatebox{90}{\textbf{Tasks}}} & \multirow{2}{*}{\textbf{LLMs/AS}} 
& \multicolumn{2}{c|}{\textbf{\ZeroCtxabbr{}}}
& \multicolumn{2}{c|}{\textbf{\PCMabbr}}
& \multicolumn{2}{c|}{\textbf{\PCMabbr+RAG}}
& \multicolumn{2}{c|}{\textbf{\PCOabbr$_5$}}
& \multicolumn{2}{c}{\textbf{\PCabbr}} \\
\cmidrule{3-12}
& & pass@1 & Pr & pass@1 & Pr & pass@1 & Pr & pass@1 & Pr & pass@1 & Pr \\
\midrule
\multirow{8}{*}{\rotatebox{90}{\textbf{Simple Tasks}}}
& GPT-4o mini   & 0.70 & 0.00 & 0.90 & 0.00 & 0.98 & 0.10 & 0.90 & 0.70 & 0.90 & 0.70 \\
& GPT-4o      & 0.70 & 0.46 & 0.90 & 0.76 & 0.98 & 0.74 & 1.00 & 0.90 & 1.00 & 0.90 \\
& GPT-5 mini    & 0.70 & 0.58 & 1.00 & 0.78 & 1.00 & 0.72 & 0.98 & 0.90 & 0.98 & 0.88 \\
& GPT-5       & 0.84 & 0.64 & 1.00 & 0.70 & 1.00 & 0.70 & \textbf{1.00} & \textbf{1.00} & \textbf{1.00} & \textbf{1.00} \\
& o4 mini       & 0.84 & 0.70 & 1.00 & 0.70 & 1.00 & 0.70 & \textbf{1.00} & \textbf{1.00} & 1.00 & 0.96 \\
& o3          & 0.82 & 0.62 & 1.00 & 0.70 & 1.00 & 0.70 & \textbf{1.00} & \textbf{1.00} & \textbf{1.00} & \textbf{1.00} \\
\cdashline{2-12} 
& Qwen2.5 7B & 0.60 & 0.00 & 0.60 & 0.00 & 0.62 & 0.00 & 0.78 & 0.48 & 0.80 & 0.20 \\
& Qwen3 8B    & 0.50 & 0.00 & 0.80 & 0.30 & 0.90 & 0.00 & 0.90 & 0.60 & 0.90 & 0.70 \\

\midrule
\multirow{8}{*}{\rotatebox{90}{\textbf{Medium Tasks}}}
& GPT-4o mini   & 0.10 & 0.00 & 0.12 & 0.00 & 0.16 & 0.06 & 0.26 & 0.06 & 0.28 & 0.16 \\
& GPT-4o      & 0.10 & 0.06 & 0.44 & 0.00 & 0.30 & 0.00 & 0.60 & 0.30 & 0.68 & 0.24 \\
& GPT-5 mini    & 0.08 & 0.00 & 0.60 & 0.00 & 0.56 & 0.00 & 0.90 & 0.36 & 0.90 & 0.28 \\
& GPT-5       & 0.20 & 0.00 & 0.82 & 0.00 & 0.72 & 0.00 & 0.78 & 0.46 & \textbf{0.98} & 0.52 \\
& o4 mini       & 0.10 & 0.02 & 0.54 & 0.04 & 0.50 & 0.02 & 0.72 & 0.44 & 0.72 & 0.46 \\
& o3          & 0.08 & 0.00 & 0.62 & 0.02 & 0.56 & 0.00 & 0.92 & 0.62 & 0.92 & \textbf{0.64} \\
\cdashline{2-12} 
& Qwen2.5 7B & 0.10 & 0.00 & 0.10 & 0.00 & 0.10 & 0.00 & 0.14 & 0.12 & 0.10 & 0.00 \\
& Qwen3 8B    & 0.10 & 0.00 & 0.10 & 0.10 & 0.10 & 0.10 & 0.20 & 0.20 & 0.20 & 0.10 \\
\midrule
\multirow{8}{*}{\rotatebox{90}{\textbf{Hard Tasks}}}
& GPT-4o mini   & 0.00 & 0.00 & 0.16 & 0.00 & 0.04 & 0.00 & 0.38 & 0.26 & 0.46 & 0.28 \\
& GPT-4o      & 0.00 & 0.00 & 0.30 & 0.00 & 0.18 & 0.00 & 0.68 & \textbf{0.52} & 0.66 & 0.34 \\
& GPT-5 mini    & 0.02 & 0.00 & 0.38 & 0.00 & 0.34 & 0.00 & 0.74 & 0.38 & 0.74 & 0.34 \\
& GPT-5       & 0.10 & 0.00 & 0.62 & 0.00 & 0.40 & 0.00 & 0.66 & 0.20 & \textbf{0.80} & 0.34 \\
& o4 mini       & 0.04 & 0.00 & 0.38 & 0.00 & 0.32 & 0.00 & 0.64 & 0.28 & 0.66 & 0.14 \\
& o3          & 0.06 & 0.00 & 0.18 & 0.00 & 0.26 & 0.00 & 0.76 & 0.48 & 0.76 & 0.46 \\
\cdashline{2-12} 
& Qwen2.5 7B & 0.00 & 0.00 & 0.00 & 0.00 & 0.00 & 0.00 & 0.12 & 0.02 & 0.00 & 0.00 \\
& Qwen3 8B    & 0.00 & 0.00 & 0.00 & 0.00 & 0.00 & 0.00 & 0.10 & 0.00 & 0.00 & 0.00 \\
\midrule
\multirow{8}{*}{\rotatebox{90}{\textbf{All Tasks}}}
& GPT-4o mini   & 0.27 & 0.00 & 0.39 & 0.00 & 0.39 & 0.05 & 0.51 & 0.34 & 0.55 & 0.38 \\
& GPT-4o      & 0.27 & 0.17 & 0.55 & 0.25 & 0.49 & 0.25 & 0.76 & 0.57 & 0.78 & 0.49 \\
& GPT-5 mini    & 0.27 & 0.19 & 0.66 & 0.26 & 0.63 & 0.24 & 0.87 & 0.55 & 0.87 & 0.50 \\
& GPT-5       & 0.38 & 0.21 & 0.81 & 0.23 & 0.71 & 0.23 & 0.81 & 0.55 & \textbf{0.93} & 0.62 \\
& o4 mini       & 0.33 & 0.24 & 0.64 & 0.25 & 0.61 & 0.24 & 0.79 & 0.57 & 0.79 & 0.52 \\
& o3          & 0.32 & 0.21 & 0.60 & 0.24 & 0.61 & 0.23 & 0.89 & \textbf{0.70} & 0.89 & 0.70 \\
\cdashline{2-12} 
& Qwen2.5 7B & 0.23 & 0.00 & 0.23 & 0.00 & 0.24 & 0.00 & 0.35 & 0.21 & 0.30 & 0.07 \\
& Qwen3 8B    & 0.20 & 0.00 & 0.30 & 0.13 & 0.33 & 0.03 & 0.40 & 0.27 & 0.37 & 0.27 \\
\bottomrule
\end{tabular}
}
\begin{minipage}{8.5cm}
\vspace{0.1cm}
\footnotesize 
$^*$\textbf{Bold} results indicate best performance across agentic systems and LLMs.
LLMs on top of the dash line are proprietary, and below are open-source.
\end{minipage}
\end{table}


\ZeroCtxabbr{} does not generalize to more complex tasks as pass@1 drops to $\text{pass@1}=0.08$–$0.20$ on medium tasks and $\text{pass@1}\leq0.10$ on hard tasks as precision stays near zero ($Pr\leq0.06$) for all proprietary models (See Table \ref{tab:avg_all}. The agent finds it hard to link multiple tools and fails the correction loop that worked on simple tasks. 
\PCMabbr{} drops sharply as task complexity increases. 
pass@1 falls to $\text{pass@1}=0.54$–$0.82$ on medium tasks and $\text{pass@1}=0.18$–$0.62$ on hard tasks, while precision remains low ($Pr\leq0.04$).
Tool descriptions that help in simple cases no longer hold when reasoning spans many tools.
\PCMabbr+RAG{} behaves just like before, showing minimal change with $\Delta P@1<0.04$ and $Pr\leq0.06$. 
While all \ZeroCtxabbr{}, \PCMabbr{}, \PCMabbr{}+RAG agentic system perform poorly, \PCOabbr$_X$ sustains $\text{pass@1}=0.72$–$0.92$ and $Pr=0.30$–$0.62$ on medium tasks and $\text{pass@1}=0.64$–$0.76$ with $Pr=0.20$–$0.52$ on hard ones. \PCabbr{} performs best with $\text{pass@1}=0.72$–$0.98$ and $Pr=0.24$–$0.64$ on medium and $\text{pass@1}=0.66$–$0.80$ and $Pr=0.14$–$0.46$ on hard. 
GPT-5 and o3 lead, reaching $\text{pass@1}=0.98$, $0.80$, and $Pr=0.64$. The open-source Qwen models follow the same trend but stay much lower ($ P@1\leq0.20 $, $ Pr\leq0.10$) even with workflow guidance.


Across all tasks (Table~\ref{tab:avg_all}), performance declines for all agentic systems as complexity increases, but \PCOabbr$_X$ and \PCabbr{} perform better than \ZeroCtxabbr{}, \PCMabbr{}, and \PCMabbr+RAG{} due to access to structured context. 
GPT-5 and o3 perform the best across the proprietary LLM models, while Qwen performs poorly with the best pass@1 metric of 0.40.


\subsection{Stress Test: Larger Set of Tools and Annotated Verified Workflow Task Pairs}
\label{sec:stress_test}

\noindent We extend the evaluation to test the scalability of the agentic systems.
In essence, we design \PCabbr{} to be generalizable and applicable for broader use and not limited to the analyses in Section~\ref{sec:analysis_tools}.
Therefore, to test the generalizability of \PCabbr{},
we expand the \ToolPool{} from 21 to 65 callable \Tools{}.
The additional 44 tools include analyses with the distribution grid besides those mentioned in Section~\ref{sec:analysis_tools}. 
Using these tools, we curate 90 new \AWF{}, bringing the total to 100.
A domain expert validates each \Workflow{} before inclusion.
We use the same evaluation protocol as given in Section~\ref{sec:metrics}.
We report aggregated results for the new expanded \ToolPool{} and \AWF{} for the same 30 tasks.
The idea is to evaluate whether \PCOabbr{} can identify the relevant workflows through a similarity search mechanism and maintain robust performance in terms of numerical correctness and workflow stability.
In addition to earlier metrics, we include pass@5 to test agentic system stability as the tool set dimension grows.
pass@1 checks single-run correctness, and pass@5 measures if at least one of the five runs gives the correct result.


From Table~\ref{tab:stress_avg_all}, pass@1 and precision remain close to the average results in Table~\ref{tab:avg_all}.
Both \PCOabbr$_X$ and \PCabbr{} sustain high pass@1 and reach pass@5 between $0.93$--$1.00$.
For \PCOabbr$_X$, pass@1 stays within $-0.04$ to $+0.07$ of the earlier value (e.g., o3: $0.89 \to 0.93$, Pr $0.70 \to 0.66$), and for \PCabbr{}, within $-0.07$ to $+0.12$ (e.g., GPT-4o-mini: $0.55 \to 0.67$, Pr $0.38 \to 0.27$).
Precision drops slightly across proprietary models, by up to $0.08$.
o3 and GPT-5 stay dominant, with o3 reaching Pr $=0.66$ under \PCOabbr$_X$ and GPT-5 achieving pass@5 $=1.00$.
Token efficiency also improves: \PCOabbr$_X$ uses 3–5$\times$ fewer tokens per pass@1 than \PCabbr{} (e.g., GPT-5: 51.3K vs.\ 205.6K, o4-mini: 47.3K vs.\ 225.7K), confirming that optimized context achieves comparable accuracy with much lower cost.
Qwen models follow the same pattern but remain far lower (e.g., \PCabbr{}: Qwen2.5 $0.30 \to 0.33$ P@1, Pr $0.07 \to 0.00$).
These patterns show that both \PCabbr{} and \PCOabbr$_X$ scale to more tools and workflows without significantly dropping in pass@1 and precision.

\begin{table}[htpb]
\centering
\footnotesize
\setlength{\tabcolsep}{4pt}
\renewcommand{\arraystretch}{1.05}
\caption{Average performance on all tasks with 65 tools and 100 annotated workflow-task pairs (10 relevant + 90 irrelevant).}
\label{tab:stress_avg_all}

\resizebox{\linewidth}{!}{
\begin{tabular}{@{}l|cccc|cccc@{}}
\toprule
\multirow{2}{*}{\textbf{LLMs/AS}} & \multicolumn{4}{c|}{\textbf{\PCOabbr$_5$}} & \multicolumn{4}{c}{\textbf{\PCabbr}} \\

\cmidrule{2-9}
& pass@1 & pass@5 & Pr & Tk/P@1 (K) & pass@1 & pass@5 & Pr & Tk/P@1 (K) \\
\midrule
GPT-4o mini & 0.47 & 0.57 & 0.33 & 15.9 & 0.67 & 0.67 & 0.27 & 98.0 \\
GPT-4o & 0.74 & 0.93 & 0.51 & 32.4 & 0.82 & 0.93 & 0.51 & 147.8 \\
GPT-5 mini & 0.87 & 0.93 & 0.47 & 50.5 & 0.87 & 0.93 & 0.47 & 229.7 \\
GPT-5 & 0.88 & \textbf{1.00} & 0.53 & 51.3 & 0.86 & 0.97 & 0.53 & 205.6 \\
o4 mini & 0.83 & 0.93 & 0.53 & 47.3 & 0.81 & \textbf{1.00} & 0.50 & 225.7 \\
o3 & \textbf{0.93} & 0.93 & \textbf{0.66} & 46.1 & 0.91 & 0.93 & 0.64 & 213.3 \\
\cdashline{1-9} 
Qwen2.5 7B & 0.42 & 0.47 & 0.30 & 27.3 & 0.33 & 0.33 & 0.00 & 102.7 \\
Qwen3 8B & 0.30 & 0.37 & 0.25 & 23.0 & 0.35 & 0.40 & 0.27 & 64.0 \\

\bottomrule
\end{tabular}
}
\end{table}

\subsection{Choice of $X$ in \PCOabbr$_X$}
\label{sec:choose_k}

\begin{figure}[htpb]
\centering
\includegraphics[width=\linewidth]{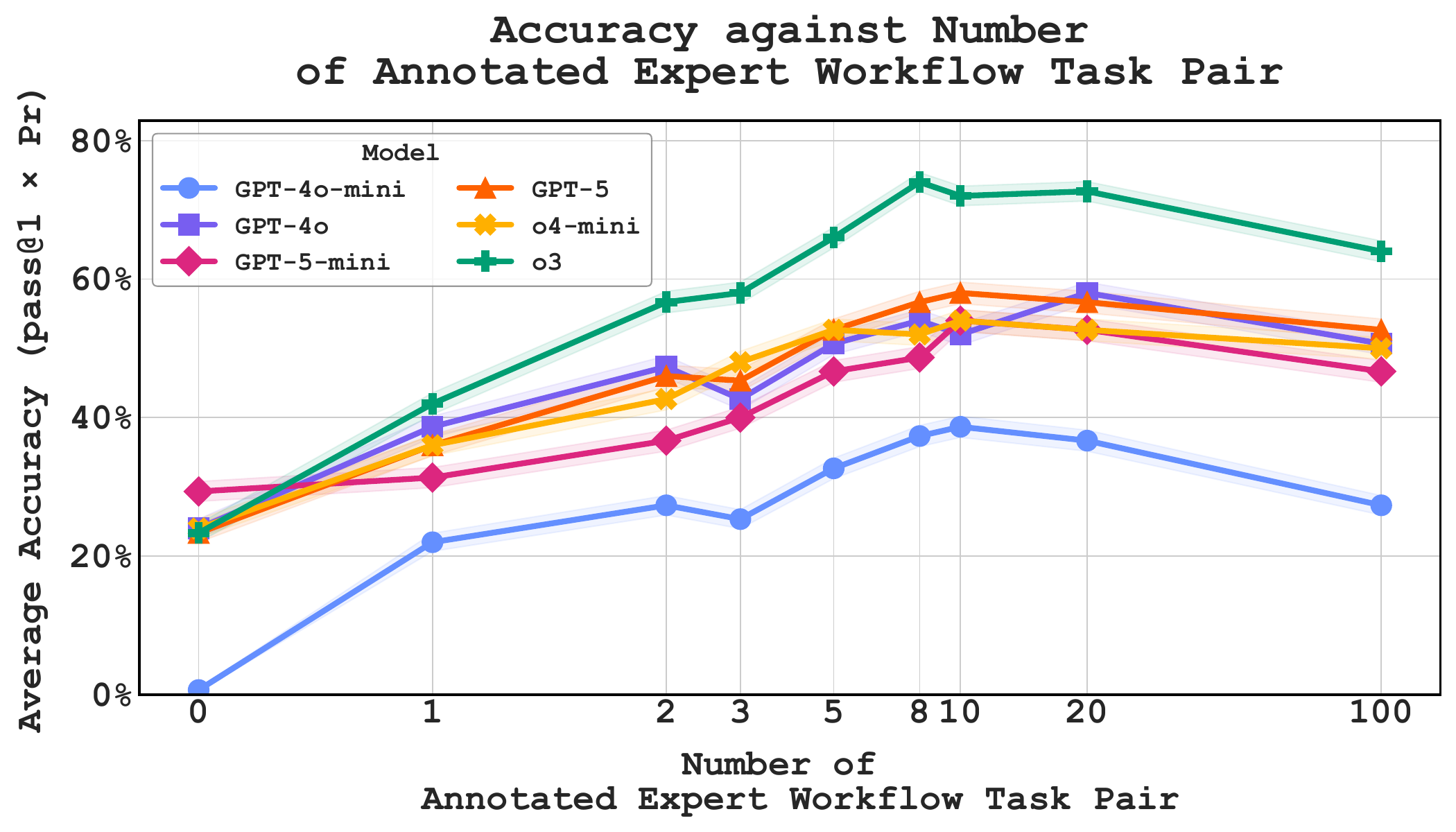}
\caption{Accuracy (pass@1~$\times$~Precision) against number of \AWF{} ($X$) for proprietary models. Accuracy increases up to $X\!\approx\!10$ and then saturates.}
\label{fig:accuracy_vs_k}
\end{figure}

\noindent The parameter $X$ specifies the number of \AWF{} incorporated into the prompt before execution.
We sweep $X\!\in\!\{1,2,3,5,8,10,20\}$ on the proprietary models to analyze how retrieval dimension impacts accuracy and token cost. 
In Fig. \ref{fig:accuracy_vs_k}, we present the impact of the choice of $X$ on accuracy, shown as the product of pass@1 and precision.
We observe a sharp rise in accuracy from $X{=}1$ to about $X{=}8$, which then plateaus until $X{=}10$, and then gradually declines as $X$ increases further, continuing to drop until we include all 100 \AWF{} in \PCabbr{}. 

We posit that the accuracy increases up to $X{=}10$ because, on closer inspection, we find that only ten \AWF{} out of the total 100 are directly relevant to the 30 unseen benchmark tasks. 
The subset selection algorithm for \AWF{} in Section~\ref{sec:optimal_awf_selection} prioritizes relevant workflows first. 
Once $X$ reaches around this number, the algorithm has included the most relevant \AWF{} and further candidates contribute little, merely increasing the prompt size and token cost. 
Larger $X$ values also introduce redundant or irrelevant \AWF{} that can affect both the pass@1 and precision score.

\begin{figure}[htpb]
\centering
\includegraphics[width=\linewidth]{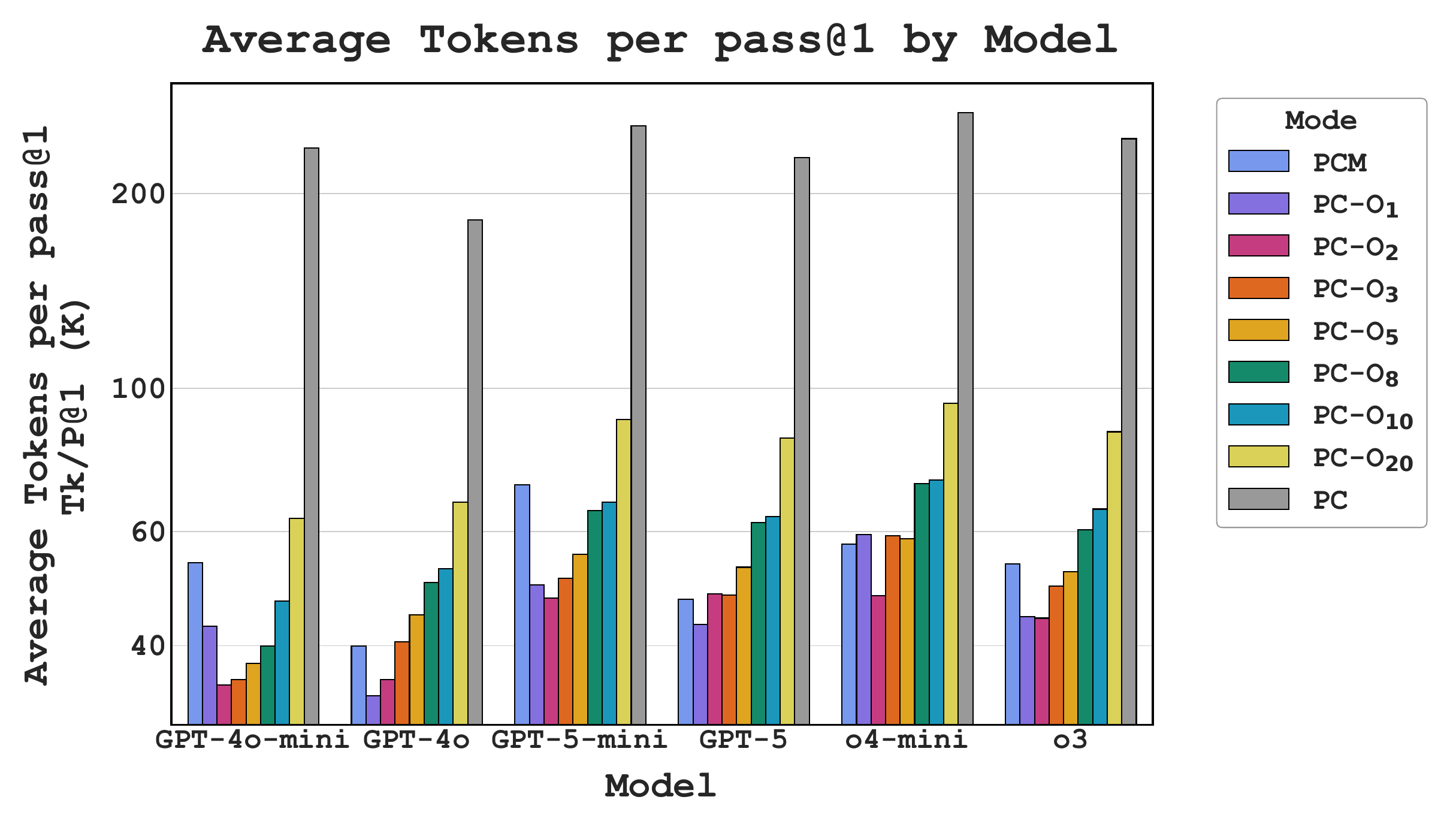}
\caption{Average tokens per pass@1 (Tk/P@1) across models and PowerChain modes.}
\label{fig:tokens_per_success}
\end{figure}

Fig.~\ref{fig:tokens_per_success} shows tokens per pass@1. The metric attains its minimum between \PCOabbr$_1$ and \PCOabbr$_2$ across the models, then increases steadily as $X$ grows.
The first one or two \AWF{} are relevant and achieve significant pass@1 growth at the expense of an increase in token cost, resulting in better accuracy than \PCMabbr{} but lower than \PCOabbr{} agentic systems with $X{=}3$–$10$.
The token use and accuracy continue to increase between \PCOabbr$_3$ and \PCOabbr$_{10}$, but token growth outpaces the accuracy (pass@1) gain, resulting in a higher token/pass@1 metric.
As $X$ grows larger than 10 in \PCOabbr{} agentic system, new \AWF{} adds orthogonal content that increases prompt size without raising accuracy (and in some instances even reduces accuracy as seen in Fig. \ref{fig:accuracy_vs_k}). 
The gap between \PCOabbr$_{10}$ and \PCabbr{} averages about $3$–$4\times$ in token cost, even though \PCabbr{} performs worse in terms of pass@1. 
When $X$ stays below or around 10, \PCOabbr$_X$ extracts most relevant \AWF{}. Beyond it, the added \AWF{} adds little structural information, increases prompt length, and reduces efficiency.

The key takeaway is that \PCOabbr$_X$ performs best when $X$ is close to the number of relevant \AWF{}, indicating that relevance coverage, rather than context size, drives both accuracy and efficiency. 
Grid-search like hyper-parameter tuning for $X$ for different sets of unseen tasks.
Still, if token cost is of utmost significance, a lower valued $X$ is desirable, but it can come at a cost of low accuracy metrics.


\section{Conclusion}
\noindent We introduce \PC{}, the first agentic AI system in power systems to orchestrate stand-alone tools to perform complex distribution grid computation tasks autonomously.
\PC{} systematically introduces structured context in LLM prompting to achieve significant gains over existing workflow automation techniques, including state-of-the-art RAG-augmented tool calling. These are the key insights we develop and empirically validate in this work:
\begin{itemize}
    \item \textit{Generalizability:} \PC{} can perform \textit{unseen} complex distribution grid compute analysis autonomously.
    \item \textit{Democratization:} \PC{} shows optimized context-aware prompting can solve complex analysis without fine-tuning LLMs.
    \item \textit{Correctability}: \PC{} embeds error-context reasoning in DAG path execution to enforce a self-correcting build–verify–feedback-correct system
\end{itemize}


\bibliographystyle{IEEEtran_custom}
\bibliography{Ref}

\begin{thebibliography}{10}
\providecommand{\url}[1]{#1}
\csname url@samestyle\endcsname
\providecommand{\newblock}{\relax}
\providecommand{\bibinfo}[2]{#2}
\providecommand{\BIBentrySTDinterwordspacing}{\spaceskip=0pt\relax}
\providecommand{\BIBentryALTinterwordstretchfactor}{4}
\providecommand{\BIBentryALTinterwordspacing}{\spaceskip=\fontdimen2\font plus
\BIBentryALTinterwordstretchfactor\fontdimen3\font minus \fontdimen4\font\relax}
\providecommand{\BIBforeignlanguage}[2]{{%
\expandafter\ifx\csname l@#1\endcsname\relax
\typeout{** WARNING: IEEEtran.bst: No hyphenation pattern has been}%
\typeout{** loaded for the language `#1'. Using the pattern for}%
\typeout{** the default language instead.}%
\else
\language=\csname l@#1\endcsname
\fi
#2}}
\providecommand{\BIBdecl}{\relax}
\BIBdecl

\bibitem{ray2023review}
S.~Ray \emph{et~al.}, ``Review of electric vehicles integration impacts in distribution networks: Placement, charging/discharging strategies, objectives and optimisation models,'' \emph{Journal of Energy Storage}, vol.~72, 2023.

\bibitem{bank2013analysis}
J.~Bank and B.~Mather, ``Analysis of the impacts of distribution connected pv using high-speed datasets,'' in \emph{2013 IEEE Green Technologies Conference (GreenTech)}.\hskip 1em plus 0.5em minus 0.4em\relax IEEE, 2013, pp. 153--159.

\bibitem{baviskar2020challenges}
A.~Baviskar \emph{et~al.}, ``Challenges of future distribution systems with a large share of variable renewable energy sources--review,'' in \emph{19th wind integration workshop}.\hskip 1em plus 0.5em minus 0.4em\relax Energynautics GmbH, 2020.

\bibitem{crozier2024distribution}
C.~Crozier \emph{et~al.}, ``Distribution grids may be a barrier to residential electrification,'' \emph{arXiv preprint arXiv:2410.04540}, 2024.

\bibitem{costello2016primer}
K.~Costello, ``A primer on r\&d in the energy utility sector,'' \emph{Energy and Environment National Regulatory Research Institute, Report}, 2016.

\bibitem{stamoulis2025geo}
D.~Stamoulis and D.~Marculescu, ``{Geo-OLM}: Enabling sustainable earth observation studies with cost-efficient open language models \& state-driven workflows,'' in \emph{ACM COMPASS}, 2025.

\bibitem{jia2024enhancing}
M.~Jia \emph{et~al.}, ``Enhancing llms for power system simulations: A feedback-driven multi-agent framework,'' \emph{IEEE Transactions on Smart Grid}, 2025.

\bibitem{lin2025novel}
H.~Lin and M.~Yu, ``A novel distributed pv power forecasting approach based on time-{LLM},'' \emph{arXiv preprint arXiv:2503.06216}, 2025.

\bibitem{jin2024chatgrid}
S.~Jin and S.~Abhyankar, ``Chatgrid: Power grid visualization empowered by a large language model,'' in \emph{2024 IEEE Workshop on Energy Data Visualization (EnergyVis)}.\hskip 1em plus 0.5em minus 0.4em\relax IEEE, 2024, pp. 12--17.

\bibitem{yang2025large}
X.~Yang \emph{et~al.}, ``Large language model powered automated modeling and optimization of active distribution network dispatch problems,'' \emph{arXiv preprint arXiv:2507.21162}, 2025.

\bibitem{andrews2025scaling}
P.~Andrews \emph{et~al.}, ``{ARE: Scaling up Agent Environments and Evaluations},'' \emph{arXiv preprint arXiv:2509.17158}, 2025.

\bibitem{zhang2025aflow}
J.~Zhang \emph{et~al.}, ``Aflow: Automating agentic workflow generation,'' in \emph{ICLR}, 2025.

\bibitem{niu2025flow}
B.~Niu \emph{et~al.}, ``Flow: Modularized agentic workflow automation,'' in \emph{The Thirteenth International Conference on Learning Representations}, 2025.

\bibitem{zhang2025multi}
G.~Zhang \emph{et~al.}, ``Multi-agent architecture search via agentic supernet,'' \emph{arXiv preprint arXiv:2502.04180}, 2025.

\bibitem{li2024autoflow}
Z.~Li \emph{et~al.}, ``{AutoFlow}: Automated workflow generation for large language model agents,'' \emph{CoRR}, 2024.

\bibitem{hu2025adas}
S.~Hu \emph{et~al.}, ``Automated design of agentic systems,'' in \emph{ICLR}, 2025.

\bibitem{su2025scaling}
L.~Su \emph{et~al.}, ``Scaling agents via continual pre-training,'' \emph{arXiv preprint arXiv:2509.13310}, 2025.

\bibitem{bhattaram2025geoflow}
A.~Bhattaram \emph{et~al.}, ``Geoflow: Agentic workflow automation for geospatial tasks,'' \emph{arXiv preprint arXiv:2508.04719}, 2025.

\bibitem{yang2025qwen3}
A.~Yang \emph{et~al.}, ``Qwen3 technical report,'' \emph{arXiv:2505.09388}, 2025.

\bibitem{openai2025o3}
OpenAI, ``Openai o3 and o4-mini system card,'' \url{https://openai.com/index/o3-o4-mini-system-card/}, 2025, accessed: Oct. 2025.

\bibitem{singh2024geollm}
S.~Singh \emph{et~al.}, ``{GeoLLM-Engine: A Realistic Environment for Building Geospatial Copilots},'' in \emph{CVPR Workshops}, 2024, pp. 585--594.

\bibitem{li2025chain}
W.~Li \emph{et~al.}, ``Chain-of-agents: End-to-end agent foundation models via multi-agent distillation and agentic rl,'' \emph{preprint arXiv:2508.13167}, 2025.

\bibitem{fang2025memp}
R.~Fang \emph{et~al.}, ``Memp: Exploring agent procedural memory,'' \emph{arXiv preprint arXiv:2508.06433}, 2025.

\bibitem{balaguer2024rag}
A.~Balaguer \emph{et~al.}, ``{RAG} vs fine-tuning: pipelines, tradeoffs, and a case study on agriculture,'' \emph{arXiv preprint arXiv:2401.08406}, 2024.

\bibitem{faiss2019}
J.~Johnson \emph{et~al.}, ``Billion-scale similarity search with gpus,'' \emph{IEEE Transactions on Big Data}, vol.~7, no.~3, pp. 535--547, 2019.

\bibitem{jia2024user}
\BIBentryALTinterwordspacing
M.~Jia \emph{et~al.}, ``User manual for {DALINE} 1.1.5: {DALINE} -- a data-driven power flow linearization toolbox for power systems research and education,'' Working Paper, ETH Zurich, 2024. [Online]. Available: \url{https://www.research-collection.ethz.ch/handle/20.500.11850/662038}
\BIBentrySTDinterwordspacing

\bibitem{matpower2024manual}
R.~D. Zimmerman and C.~E. Murillo-S{\'a}nchez, \emph{MATPOWER 8.0 User's Manual}, 2024, online. Available: \url{https://matpower.org/docs/MATPOWER-manual-8.0.pdf}.

\bibitem{misrahi2025adapting}
A.~Misrahi \emph{et~al.}, ``Adapting large language models for multi-domain retrieval-augmented-generation,'' \emph{arXiv preprint arXiv:2504.02411}, 2025.

\bibitem{zeng2025worse}
L.~Zeng \emph{et~al.}, ``Worse than zero-shot? a fact-checking dataset for evaluating the robustness of {RAG} against misleading retrievals,'' in \emph{Advances in Neural Information Processing Systems (NeurIPS)}, 2025.

\bibitem{cheng2025large}
Y.~Cheng \emph{et~al.}, ``A large language model for advanced power dispatch,'' \emph{Scientific Reports}, vol.~15, no.~1, p. 8925, 2025.

\bibitem{gao2025beyond}
J.~Gao \emph{et~al.}, ``Beyond ten turns: Unlocking long-horizon agentic search with large-scale asynchronous rl,'' \emph{arXiv preprint arXiv:2508.07976}, 2025.

\bibitem{openai2025deepresearch}
OpenAI, ``Introducing deep research,'' \url{https://openai.com/index/introducing-d eep-research/}, 2025, accessed: Oct. 2025.

\bibitem{jimenezswe}
C.~E. Jimenez \emph{et~al.}, ``{SWE-bench}: Can language models resolve real-world github issues?'' in \emph{ICLR}, 2024.

\bibitem{yao2024tau}
S.~Yao \emph{et~al.}, ``$\tau$-bench: A benchmark for tool-agent-user interaction in real-world domains,'' \emph{arXiv preprint arXiv:2406.12045}, 2024.

\bibitem{patil2025bfcl}
S.~G. Patil \emph{et~al.}, ``The berkeley function calling leaderboard ({BFCL}): From tool use to agentic evaluation of large language models,'' in \emph{Forty-second International Conference on Machine Learning}, 2025.

\bibitem{michelakis2025core}
P.~Michelakis \emph{et~al.}, ``{CORE}: Full-path evaluation of {LLM} agents beyond final state,'' in \emph{LAW NeurIPS Workshop}, 2025.

\bibitem{foster2022three}
E.~Foster \emph{et~al.}, ``Three-phase infeasibility analysis for distribution grid studies,'' \emph{Electric Power Systems Research}, vol. 212, p. 108486, 2022.

\bibitem{pandey2018robust}
A.~Pandey \emph{et~al.}, ``Robust power flow and three-phase power flow analyses,'' \emph{IEEE Transactions on Power Systems}, vol.~34, 2018.

\bibitem{badmus2024anoca}
E.~O. Badmus and A.~Pandey, ``{ANOCA}: {AC} network-aware optimal curtailment approach for dynamic hosting capacity,'' in \emph{IEEE 63rd Conference on Decision and Control (CDC)}.\hskip 1em plus 0.5em minus 0.4em\relax IEEE, 2024.

\bibitem{panthee2025solving}
B.~Panthee and A.~Pandey, ``Solving three-phase {AC} infeasibility analysis to near-zero optimality gap,'' \emph{arXiv preprint arXiv:2508.15937}, 2025.

\bibitem{sclar2023quantifying}
M.~Sclar \emph{et~al.}, ``Quantifying language models' sensitivity to spurious features in prompt design or: How i learned to start worrying about prompt formatting,'' \emph{arXiv preprint arXiv:2310.11324}, 2023.

\bibitem{TRIPATHI2025841}
S.~Tripathi \emph{et~al.}, ``A hitchhiker's guide to good prompting practices for large language models in radiology,'' \emph{Journal of the American College of Radiology}, vol.~22, no.~7, pp. 841--847, 2025.

\bibitem{wang2022text}
L.~Wang \emph{et~al.}, ``Text embeddings by weakly-supervised contrastive pre-training,'' \emph{arXiv preprint arXiv:2212.03533}, 2022.

\bibitem{liu2022using}
M.~Z. Liu \emph{et~al.}, ``Using {OPF}-based operating envelopes to facilitate residential {DER} services,'' \emph{IEEE Transactions on Smart Grid}, vol.~13, no.~6, pp. 4494--4504, 2022.

\bibitem{moring2023inexactness}
H.~Moring and J.~L. Mathieu, ``Inexactness of second order cone relaxations for calculating operating envelopes,'' in \emph{2023 IEEE International Conference on Communications, Control, and Computing Technologies for Smart Grids (SmartGridComm)}.\hskip 1em plus 0.5em minus 0.4em\relax IEEE, 2023, pp. 1--6.

\bibitem{yi2022fair}
Y.~Yi and G.~Verbi{\v{c}}, ``Fair operating envelopes under uncertainty using chance constrained optimal power flow,'' \emph{Electric Power Systems Research}, vol. 213, p. 108465, 2022.

\bibitem{langchain2024}
LangChain, ``Integrations: Faiss — usage for retrieval-augmented generation,'' \url{https://python.langchain.com/docs/integrations/vectorstores/faiss/}, 2024, accessed: Oct 2025.

\bibitem{openai2024embeddings}
{OpenAI}, ``Text embedding models: \texttt{text-embedding-3-small} and \texttt{text-embedding-3-large},'' \url{https://platform.openai.com/docs/guides/embeddings}, 2024, accessed: Oct 2025.

\bibitem{koh2024visualwebarena}
J.~Y. Koh \emph{et~al.}, ``{VisualWebArena}: Evaluating multimodal agents on realistic visual web tasks,'' in \emph{{ICLR} Workshop on {LLM} Agents}, 2024.

\bibitem{zhou2024webarena}
S.~Zhou \emph{et~al.}, ``{WebArena}: A realistic web environment for building autonomous agents,'' in \emph{ICLR}, 2024.

\bibitem{chen2021evaluating}
M.~Chen \emph{et~al.}, ``Evaluating large language models trained on code,'' \emph{arXiv preprint arXiv:2107.03374}, 2021.

\end{thebibliography}

\scriptsize
\setlist[enumerate]{itemsep=0pt, parsep=0pt, topsep=3pt, partopsep=0pt, leftmargin=*, label=\arabic*.}

\begin{appendices}
\section{Benchmark Task Set}
\label{appx:queries}

\noindent
We show representative samples from the 30  benchmark tasks used for PowerChain evaluation. 
Each task is phrased as a natural-language query that the agent must translate into executable workflows. 

\subsection*{Easy}
\begin{enumerate}
    \item How many capacitors are in the Rochester feeder, Vermont?
    \item[] \centerline{\vdots}
    \setcounter{enumi}{9}
    \item Can you export all nodes whose voltage is 120 volts in the South Hero feeder into a txt file?
\end{enumerate}

\subsection*{Medium}
\begin{enumerate}[resume]
    \item Without running any analysis, plot the bus voltage magnitudes from the input data for the Glover distribution network at 11:00 AM, 18th of March, 2025.
    \item[] \centerline{\vdots}
    \setcounter{enumi}{19}
    \item For the Glover network at 6:00 PM on March 21, 2025, run L1 current infeasibility analysis. Set the convergence tolerance to 1e-6. Report the top 5 buses with the highest infeasible currents. Then, plot the voltage magnitude after this analysis.
\end{enumerate}

\subsection*{Hard}
\begin{enumerate}[resume]
    \item Load network file for the Rochester feeder in Vermont on 2025-03-19 at 15:00. Run dynamic hosting capacity. Assume solar irradiance data for that date and time. Choose a sparse curtailment strategy. Enforce the voltage limit between 0.95 and 1.05 at all nodes. Limit transformer flow to 110\% of rated. After solving the problem, plot the curtailed power as a feeder topology map, with each bus color-coded by its curtailed power value and edges drawn according to the network connectivity.
    \item[] \centerline{\vdots}
    \setcounter{enumi}{29}
    \item Conduct a L2 norm current infeasibility study for the Stowe feeder, representing conditions at 10:00 on March 23, 2025. In this analysis, a global voltage of 0.94–1.06 p.u. is applied to all the buses except buses with IDs of 10, 25, and 36. These buses' voltage should be between 0.98 and 1.02 p.u. Additionally, enforce a global transformer loading limit of 115\% of its rating. Solve the model using IPOPT, with a tolerance of 1e-6 and a maximum iteration count of 1000. After solving the problem, generate a plot visualizing the infeasible currents of each bus on a map, and another plot showing the bus voltage on a map.
\end{enumerate}


\section{Annotated Verified Workflow-Task Pairs}
\label{appx:awf}

\noindent
This appendix shows the ten annotated verified workflow–query pairs (\AWF{}) used as expert exemplars during model training and evaluation. 
Each pair links a natural-language task to a verified domain workflow. 

\begin{enumerate}
    \item Load the \{location\} feeder model and report counts of buses, lines, and transformers as a quick integrity check.
    \item[] \centerline{\vdots}
    \setcounter{enumi}{9}
    \item For the \{location\} feeder at \{YYYY-MM-DD\} \{HH:MM\}, run L2 current-infeasibility with transformer limits at 110\%, applying global voltage bounds of 0.90–1.10 p.u. and custom bounds of 0.95–1.05 p.u. for 5th, 10th, and 30th buses by ID. Then plot both the bus voltages and infeasible currents.
\end{enumerate}

\noindent Full benchmark task set and complete \AWF{} repository are available at: \url{https://github.com/emmanuelbadmus/powerchain}

\noindent We note that while this work is limited to ten representative queries, ongoing collaborations with VEC and other partners aim to scale the benchmark to hundreds of expert-level tasks. We also plan to open-source additional datasets and agent baselines to support reproducible power-grid analytics research.

\normalsize
\end{appendices}

\end{document}